\documentclass{article}
\usepackage{microtype}
\usepackage{graphicx}
\usepackage{subfigure}
\usepackage{booktabs} 

\usepackage{hyperref}
\usepackage{latexsym,amsmath,amsfonts,amsthm,amssymb,mathtools}

\usepackage{algorithm}
\usepackage{algorithmic}

\newcommand{\R}{\mathbb R}
\newcommand{\E}{\mathbb E}
\newcommand{\Gcal}{\mathcal G}
\newcommand{\Vcal}{\mathcal V}
\newcommand{\Mcal}{\mathcal M}

\newtheorem{remark}{Remark}
\newtheorem{theorem}{Theorem}

\begin{document}
\title{Stochastic 
	Runge-Kutta methods and adaptive SGD-G2 stochastic gradient descent}

\author{Imen AYADI and Gabriel TURINICI \\
	CEREMADE, Université Paris Dauphine - PSL Research University, Paris, France\\
\\ correspondence to 
	\texttt{Gabriel.Turinici@dauphine.fr}}
\date{February 9, 2020}
\maketitle 

\begin{abstract}
The minimization of the loss function is of paramount importance in deep neural networks. On the other hand, many popular optimization algorithms have been shown to correspond to some evolution equation of gradient flow type. Inspired by the numerical schemes used for general evolution equations we introduce a second order stochastic Runge Kutta method and show that it yields a consistent procedure for the minimization of the loss function. In addition it can be coupled, in an adaptive framework, with a Stochastic Gradient Descent (SGD) to adjust automatically the learning rate of the SGD, without the need of any additional information on the Hessian of the loss functional.
The adaptive SGD, called SGD-G2, is successfully tested on standard datasets.

\end{abstract}

\section{Introduction and related literature}

Optimization algorithms are at the heart of neural network design in deep learning. One of the most studied procedures is the fixed step Stochastic Gradient Descent (SGD) \cite{bottou_stochastic_2012}; although very robust,  SGD may converge too slow for small learning rates or become unstable if the learning rate is too large. Each problem having its own optimal learning rate, there is no general recipe to adapt it automatically; to address this issue, several approaches have been put forward among which the use of momentum \cite{sutskever_importance_2013}, Adam \cite{kingma2014adam}, RMS\-prop \cite{tieleman_lecture_2012} and so on. On the other hand, recent research efforts have been directed towards finding, heuristically or theoretically, the best learning rate \cite{seong_towards_2018,smith_super-convergence_2017,smith_cyclical_2017} with \cite{yedida_lipschitzlr_2019}, which uses an estimate of the Lipschitz constant, being a very recent example.

Another interpretation of the minimization procedure it to see it as a time evolution (a flow) in the space $\R^d$ of neural network parameters. Denote such a parameter by $X$ and let 
$X \in \R^d \mapsto f(X) \in \R$ be the 
the loss functional; the flow interpretations recognizes that the minimization of $f(X)$ is related to the solution of the following evolution equation 
\begin{equation}
    X'(t) =  \nabla f(X(t)).
\label{eq:evolution}
\end{equation}

In this work we consider the flow \eqref{eq:evolution} and apply two  numerical schemes to evolve it in time: the 
Explicit Euler scheme (which will correspond to the SGD algorithm) and a numerical scheme of second order in time, labeled "SH" (like in "stochastic Heun") belonging to the class of stochastic Runge-Kutta methods; this second scheme allows to have a more precise estimation of the flow and in turn 
provides essential information to adapt the learning rate of the SGD.

We prove theoretically in section \ref{sec:theory} that the SH scheme is indeed of second order; then we explain how it allows to choose the optimal learning rate for the SGD and build the SGD-G2 algorithm. Numerical results on standard datasets (MNIST, F-MNIST, CIFAR10) are presented in section \ref{sec:numerics} followed by a discussion and concluding remarks.

\section{Notations}
\label{sec:notations}

The fit of the neural networks is formalized 
through the introduction of a loss functional 
$f$ depending on the network parameters $X$
and the input data $\omega \in \Omega$ presented to it. In full generality the input data belongs to some
probability space $(\Omega, {\mathbb P})$ and the optimization aims to find the value $X^{opt}$ minimizing the mapping
$X \mapsto \E_\omega f(\omega,X)$. 
However, for instance for classification purposes, not all $\omega$ have a label attached to it, so 
in practice only a limited amount of values $\omega_1,... , \omega_N \in \Omega$ can be used. So
the loss functional becomes:
\begin{equation}
    f(X) : = \frac{1}{N} \sum _{i=1}^{N}f(\omega_i,X).
\end{equation}
 For $i \le N$ we introduce the functions $f_i = f(\omega_i,\cdot) : \R^d \to \R$ to represent the loss due to the $i^{th}$ training sample $\omega_i$.

To  minimize 
the loss functional one can think of an deterministic procedure (of gradient descent type) which can be  written as: 
\begin{equation}
    X_{n+1} = X_n - h \nabla f(X_n),
\end{equation}
where $h > 0$ denotes the learning rate (also called "step size").
Note that this update rule requires $N$ gradient evaluations per step which is prohibitively large in applications in 
deep learning that involve networks with many parameters (tenths of thousands up to billions). To this end, the deterministic procedure 
is replaced by its stochastic counterpart, the Stochastic Gradient Descent (SGD).
Let $(\gamma_n)_{n \ge 1 } $ be i.i.d uniform variables taking values in $\left\{1,2,..,N\right\}$. Then, the SGD  is defined as:
\begin{equation}
    X_{n+1} = X_n - h \nabla f_{\gamma_n}(X_n), \ X_0=X(0). \label{eq:sgddef}
\end{equation}
The advantage of the stochastic algorithm is that the gradient is evaluated once per iteration which makes its computational complexity independent of $N$. This explains why this method is preferred for large data sets.\footnote{In practice $\gamma_n$ are drawn without replacement from $\{1, ..., N\}$ until all values are seen. This is called an epoch. Then all values re-enter the choice set and a new epoch begins. We will not take discuss this refinement in our procedures which is independent of the segmentation in epochs or not. Same for mini-batch processing which consists in drawing several $\gamma_k$ at once; the method proposed in the sequel adapts out-of-the-box to such a situation too.}

\subsection{The construction of the SGD-G2 algorithm: the principle}

The state $X_{n+1}$ in \eqref{eq:sgddef} can be also seen as an approximation of the solution $X(t)$ of the flow in \eqref{eq:evolution} at the "time" $t_{n+1}=(n+1)h$:
 $X_{n+1} \simeq X(t_{n+1})$.
 But there are many ways to obtain approximations of 
$X(t_{n+1})$, for instance one can use a second order 
in $h$ scheme (such as the so-called Runge-Kutta schemes to name but a few \cite{press_numerical_2007}) and construct another 
approximation $Y_{n+1}$ at the price of computing another gradient. If $Y_{n+1}$ is a better approximation then it closer to 
$ X(t_{n+1})$ and thus at the leading order 
$Y_{n+1}-X_{n+1}$ is an estimation of the error
$X_{n+1}-X(t_{n+1})$. With such an approximation one can extrapolate the behavior of $f$ near $X_n$ and compute for what values of the learning rate $h$ we still have stability (the precise computations are detailed in the next section). We adapt then the learning rate to go towards the optimal value, that is, large enough to advance fast but still stable. This will be encoded in the SGD-G2 algorithm we propose.
Two questions arise:

- how to design a high order scheme consistent with the equation \eqref{eq:evolution}: this is the object of section \ref{sec:theory};

- is the numerical procedure performing well in practice : this is the object of section \ref{sec:numerics}.

\section {Theoretical results} \label{sec:theory}

\subsection{Choice of the stochastic Runge-Kutta scheme}

First we need to choose a numerical scheme that solves the equation \eqref{eq:evolution} by using only partial information on the samples, i.e., we have to use some stochastic numerical scheme. Among the possible variants we choose the stochastic-Heun method described below (see also \cite{ruthotto_deep_2018,zhu_convolutional_2019} for related works, although not with the same goal); while the SGD updates the state by relation \eqref{eq:sgddef} the stochastic Heun scheme (named "SH" from now on) reads:
\begin{eqnarray}
    & \ & Y_0 = X(0)
    \nonumber \\  
    & \ & \tilde{Y}_{n+1} = Y_n - h \nabla f_{\gamma_n}(Y_n) 
    \nonumber \\ & \ & 
    Y_{n+1} = Y_n - \frac{h}{2} 
    \left[
    \nabla f_{\gamma_n}(Y_n)
    +\nabla f_{\gamma_n}(\tilde{Y}_{n+1}) \right]
    \label{eq:sto-heun_def}
\end{eqnarray}
    Note that a step requires two evaluations of the gradient, but this is the price to pay for higher precision. 
    Note also that the same random sample $\gamma_n$ is used
    for both gradient computations.
    In fact SH this can be also seen as a SGD 
that uses a sample {\bf twice} to advance and then
do a linear combination of the gradients thus obtained.

In order to prove relevant properties of SH scheme, we need to make clear some details concerning the evolution equation \eqref{eq:evolution}. In fact, since only one sample is used at the time, the evolution $X_n$ will depend on the order in which samples $\omega_{\gamma_n}$ are chosen. This means that in fact there is some randomness involved and we cannot hope to approach exactly the solution $X(t)$ of \eqref{eq:evolution}. In fact, see 
\cite{weinan_e_sgd_as_sde_2017,li_stochastic_2019}, it is known that the output of, let's say, the SGD algorithm is close 
(in mathematical terms "weakly converging") when $h\to 0$ to the solution of the following stochastic differential  equation:
\begin{equation}
    d Z_t = b(Z_t) dt + \sigma(Z_t) d W_t, \ Z(0) = X(0),
    \label{eq:sdegeneral}
\end{equation}
with $b(z) = - \nabla f(z)$ and 
$\sigma(z) = (h \Vcal(z))^{1/2}$, where 
\begin{equation}
    \Vcal(z) =  \frac{ \sum_{k=1}^{N} (\nabla f_{\gamma_k}(z) - \nabla f(z))(\nabla f_{\gamma_k}(z) - \nabla f(z))^T}{N},
\end{equation}
is the covariance matrix of $\nabla_X f(\omega,z)$ taken as a random variable of the samples $\omega$
(see also \cite{weinan_e_sgd_as_sde_2017} equation (4)). Here $W_t$ is a standard Brownian motion. 
With these provisions we can formally state the following result:
\begin{theorem}[Convergence of SGD and SH schemes] Suppose $f$, $f_k$ are Lipschitz functions having at most linear increase for $|X| \to \infty$\footnote{The linear growth at infinity is a technical hypothesis. It is for instance true when the parameter domain is closed and bounded and the function continuous.}.
The SGD scheme converges at (weak) order $1$ (in $h$) to the solution $Z_t$ of \eqref{eq:sdegeneral} while 
the SH scheme \eqref{eq:sto-heun_def} at (weak) order $2$.
\label{thm:cvsgdrk2}
\end{theorem}

\noindent {\bf Proof:} Te recall what weak convergence 
means we need to introduce some notations: we designate by $\Gcal$ the set of function having at most polynomial growth at infinity
and $W^{1,\infty}$ the set of Lipschitz functions. Given a numerical scheme $U_n$ of step $h$ (SGD, SH, etc.) that approximates the solution $Z_t$ of the SDE \eqref{eq:sdegeneral} with $U_n \simeq Z_{n h}$, weak convergence at order $p$ means that, for any 
 $t$ (kept fixed) 
 and any function $G \in \Gcal$ 
 we have\footnote{We used the notation $\lfloor x \rfloor$ to designate the integer part of a real number $x$, that is the largest integer smaller than $x$.} :
 \begin{equation}
|\E G(U_{\lfloor t/n \rfloor}) - \E G(Z_t)| = O(h^p).
 \end{equation}

It is known from \cite{weinan_e_sgd_as_sde_2017} (Theorem 1 point "i") that SGD is of weak order $1$. It remains to
 prove that SH is of weak order $2$; the proof uses Theorem 2 from the same reference (see also \cite{milshtein_weak_1986}) that is recalled below:
\begin{theorem}[Milstein,1986]
 Suppose $\forall i\ge 1$ : $\nabla f, \nabla f_i \in 
    \Gcal \cap W^{1,\infty},$   and have  at most linear growth at infinity. Suppose that $Z_0 =U_0 = z \in \R^d$.
    Let $\Delta = (\Delta_1, ..., \Delta_d) = Z_h -z$ and $\overline{\Delta} = 
    (\overline{\Delta}_1, ..., \overline{\Delta}_d)=
    U_1 - z$. If in addition, there exist $K_1, K_2 \in \Gcal$ such that for any
    $ s \in \{ 1,2,...,2 p+1\}$ and any $z$:
    \begin{equation}
             \left| \E(\  \prod_{j=1}^{s}\Delta_{i_j} \ ) - \E(\  \prod_{j=1}^{s} \overline{\Delta}_{i_j}\ ) \right| 
             \leq K_1(x) \ h^{p+1},
    \end{equation}
    and 
    \begin{equation}
             \E(\ \prod_{j=1}^{2p+1} |\overline{\Delta}_{i_j}| ) \leq K_2(x) \ h^{p+1},
    \end{equation}
    the numerical scheme $U_n$ converges at weak order $p$.
    \label{thm:milstein}
    \end{theorem}


We return to the proof of theorem \ref{thm:cvsgdrk2}. 
Let $L$ be an operator acting over sufficiently smooth functions 
$\zeta  :\R^d \to \R$ by:
\begin{equation}
    L\zeta = - <\nabla f,\nabla \zeta> + \frac{h}{2} \sum_{i,j=1}^{d} \Vcal_{ij} \partial^2_{ij} \zeta.
\end{equation}

Let $\Phi \in \Gcal$ and suppose it is $6$ times differentiable; using a classical result of semi-groups expansions (see \cite{phillips_functional_1996}), we have:
\begin{equation}
    \E(\Phi(Z_h)) = \Phi(z) + h L\Phi(z) + \frac{h^2}{2} 
    L^2\Phi(z) + O(h^3).
\label{eq:semigroupexpansion}
\end{equation}
For $\xi \in \R^d$, we define  $\Phi_\xi : y \mapsto e^{i \langle \xi , y-z \rangle}$.
Let $\Mcal : \R^d \to \R$ defined by $\Mcal(\xi)= e^{i< \xi,\Delta>}$.
Note that the function $\Mcal$ belongs to the class 
$ {\mathcal {C}}^{\infty}(\R^d)$ and for $s=1,..,d$, 
\begin {equation}
    \left. \frac{\partial^s \Mcal}{\prod_{j=1}^{s}\partial \xi_{i_j}}\right| _{\ \xi=0} = i^s   \E(\prod_{j=1}^{s} \Delta_{i_j}). 
\end{equation}
To determine the partial derivatives of $\Mcal$ in $0$ , we use 
\eqref{eq:semigroupexpansion} to get:
\begin{equation}
    \Mcal(\xi) = 1 + h L\Phi_t(z) +  \frac{h^2}{2}  L^2\Phi_\xi(z) + O(h^3).
\end{equation}
After calculating the explicit expressions of $L\Phi_\xi(z)$ and $L^2\Phi_\xi(z)$, we obtain:
\begin{eqnarray}
& \ & 
    \Mcal(\xi) = 1 - i h  \langle \nabla f(z),\xi \rangle 
    \nonumber \\ & \ & 
    \!\!\!\!\!\!\!\!    \!\!\!\!\!\!\!\!
    + h^2 \left( \frac{i}{2} \sum_{k=1}^{d} \partial_k f(z)  
    \langle \partial_k \nabla f(z),\xi \rangle 
    + \langle \nabla f(z),\xi \rangle^2 
    -\frac{1}{2} \xi^T \Vcal \xi \right)
        \nonumber \\ & \ & 
    \!\!\!\!\!\!\!
    + O(h^3) .
\end{eqnarray}    
Therefore,
\begin{equation} 
    \E(\Delta_j) = - h \partial_j f(z) + 
    \frac{h^2}{2} \sum_{k=1}^{d} \partial_k f(z) 
    \partial^2_{k j} f(z) + O(h^3) 
    \end{equation}
\begin{equation} 
    \E(\Delta_j \Delta_l) = h^2 [\partial_j f(z)\partial_l f(z) + \Vcal_{jl} ] +  O(h^3)
\end{equation}
\begin {equation} 
    \E(\prod_{j=1}^{s} \Delta_{i_j}) = O(h^3), \  for \  s \ge 3. \end{equation}
For the SH scheme:
\begin{equation}
    E(\overline{\Delta}_k) = -h \partial_k f(z) - \frac{h^2}{2} <\partial_i \nabla f(z),\nabla f(z)>+ O(h^3),
\end{equation}
\begin{equation}
    \E(\overline{\Delta}_k \overline{\Delta}_l) = \frac{1}{N} h^2 \sum_{\ell=1}^{N} \partial_k f_\ell(z)   \partial_l f_\ell(z) + O(h^3),
\end{equation}
\begin{equation}
    \E(\prod_{j=1}^{s} \overline{\Delta}_{i_j}) = O(h^3),
    \ \textrm{for }  s \geq 3.
\end{equation}
Therefore, all hypotheses of the theorem \ref{thm:milstein}
are satisfied for $p=2$ which gives the conclusion.

Given theorem \ref{thm:cvsgdrk2} we can trust SH to produce high order approximations of the solution which in turn will help obtain information on the Hessian and thus calibrate automatically the learning rate.

\subsection{Rationale for the adaptive step proposal}

In what follows, $ \langle X,Y \rangle $ denotes 
the usual scalar product in $\R^d$ and
$\|X\|$ the associated euclidean norm.
For a matrix $A \in \R^{d \times d}$, 
$\|A\|$ denotes the matrix norm defined as: 
$\|A\|= \sup_{X \in \R^d, X \neq 0} 
\frac {\|AX||}{\|X\|}$.

If the loss function $f$ is smooth enough, a Taylor expansion around a current point $Y$ allows to write:
\begin{eqnarray}
& \ & 
\!\!\!\!\!\!\!\!\!\!\!\!\!\!\!
f(X) = f(Y) + \langle \nabla f(Y), X-Y \rangle 
\nonumber \\ & \ &  
\!\!\!\!\!\!\!\!\!\!\!\!
+\frac{1}{2} \langle \nabla^2 f(Y) ( X-Y), X-Y \rangle
+ O(\| X-Y \|^3),
\end{eqnarray}
where $\nabla f$ is the gradient of $f$ computed at $Y$ and  $\nabla^2 f(Y)$ is the Hessian matrix
of $f$ at the same point. 

Note that the Hessian provides detailed information over the behavior of $f$ around the current point $Y$ but in practice $\nabla^2 f(Y)$  is a high dimensional object and it is impossible to deal with it directly.
We will not try to compute it but will exploit its structure.

Neglecting higher order terms, the loss functional can be written as:
 \begin{equation}
f(X) \simeq \frac{1}{2} \langle AX, X \rangle  - 
\langle b,X\rangle,
\label{eq:forder2}
 \end{equation}
 for some matrix $A$ in $\R^{d \times d}$ 
 and vector $b$ in $\R^d$.
To explain our method we will consider 
that in equation \eqref{eq:forder2} we have equality.
Then, $\nabla f(X) = AX - b $.
If $X^{opt}\in \R^d$ is the minimum of $f$, then 
$A X^{opt} = b$ and thus 
\begin{equation}
    \nabla f(X) = A(X-X^{opt}).
\label{eq:gradientformula}
\end{equation}

We will forget for a moment that the gradient of $f$ is not computed exactly (only an unbiased estimator being available in practice under the form of $\nabla f_\gamma$).
To minimize the function $f$, the
gradient descent (also called Explicit Euler) scheme
with learning rate $h_n >0$ reads:
\begin{equation}
X_{n+1} = X_n - h_n \nabla f(X_n).
\label{eq:defxrecurence}
\end{equation}
Then, denoting $I_d$ the identity matrix of $d$ dimensions, we obtain:
\begin {equation}
    X_{n+1} - X^{opt} = (I_d - h_n A) (X_n - X^{opt}).
\label{eq:defx}
\end {equation}
On the other hand the Heun scheme (which is a Runge Kutta method of the second order) reads:
\begin{equation}
   Y_{n+1} = Y_n - \frac{h_n}{2} ( \nabla f(Y_n) + \nabla f(Y_n - h_n \nabla f(Y_n))).
\label{eq:defyrecurence}
\end{equation}
Then, 
\begin{equation}
    Y_{n+1} - X^{opt} =  \left[ 
    I_d - h A + \frac{h_n^2 A^2}{2} \right] (Y_n - X^{opt}).
\label{eq:defy}
\end{equation}
At the step $n$, suppose that the two schemes start from the same point i.e., $X_n$ = $Y_n$. Therefore, using \eqref{eq:defx} and \eqref{eq:defy}:
\begin{equation}
    Y_{n+1} - X_{n+1} = \frac{h_n^2 A^2}{2}(X_n - X^{opt}) 
    = \frac {h_n^2 A}{2} \nabla f(X_n)
\label{eq:xy1}
\end{equation}
On the other hand from \eqref{eq:defxrecurence} 
and \eqref{eq:defyrecurence} we can also write:
\begin{equation}
    Y_{n+1} - X_{n+1} = \frac{h_n}{2} (\nabla f(X_n) - \nabla f(X_{n+1}))
\label{eq:xy2}
\end{equation}
Combining \eqref{eq:xy1} and \eqref{eq:xy2} we get:
\begin{equation}
    (I_d - h_nA) \nabla f(X_n) = \nabla f(X_{n+1}).
\end{equation}
From \eqref{eq:gradientformula}
and \eqref{eq:defx} the stability criterion 
for the gradient descent (Explicit Euler) scheme is
that $\| \nabla f(X_n)  \|$ needs to be bounded, 
which is verified 
in particular when
 $\| I_d - h_nA \| < 1$. If this condition is true 
 then for each step $n$:
\begin{equation}
    \frac {\|(I_d - h_nA) \nabla f(X_n)\|}{\|\nabla f(X_n)\|} < 1.
\end{equation}
Although the reciprocal is false, it is not far from true because, if $\frac {\|(I_d - h_nA) \nabla f(X_n)\|}{\|\nabla f(X_n)\|} > 1$  for some $n$, then this means in particular 
$\|(I_d - h_nA) \| > 1$ (because the matricial norm is a supremum) and then, except degenerate initial conditions $X_0$,  we obtain that $X_n$ will diverge (unless $h_n$ is adapted to ensure stability).
So to enforce stability we have to request: 
\begin{equation}
     \|\nabla f(X_{n+1})\| \le \| \nabla f(X_n)\|.
\label{eq:stabilitycondition}
\end{equation}
On the other hand, at every iteration $n$, 
we attempt to choose the biggest possible learning rate that guarantees \eqref{eq:stabilitycondition}, that is we accelerate the rate of convergence without breaking the stability criterion. 

A natural question arises : what is the maximum
value of $h$ so that \eqref{eq:stabilitycondition} still holds ?

Let us denote $\xi_n(h) = ||(I_d - h A) \nabla f(X_n)||^2$. With $X_n$ being given, this is a second order polynomial in $h$; let $h^{opt}_n$ be the maximum value of the learning rate $h$ such \eqref{eq:stabilitycondition} still holds. 
In other words, $\xi_n(h^{opt}_n) = \|\nabla f(X_n) \|^2 $.

Note that:
\begin{eqnarray}& \ & 
\xi_n(h) = \|A \nabla f(X_n)\|^2 h^2 
\nonumber \\ & \ & 
-2  \langle A\nabla f(X_n),\nabla f(X_n) \rangle h 
+ \|\nabla f(X_n)\|^2.
\end{eqnarray}

Then, $\xi_n(h^{opt}_n)=||\nabla f(X_n)||^2$ implies that $h^{opt}_n = 0$ or $h^{opt}_n = \frac{2  \langle 
A\nabla f(X_n),\nabla f(X_n) \rangle }{\| A \nabla  f(X_n)\|^2} $.
Since $(I_d - h_n A) \nabla f(X_n) = \nabla f(X_{n+1})$, we have that $A\nabla f(X_n) = \frac{ \nabla f(X_n) - \nabla f(X_{n+1})}{h_n}$. Then, unless  
$\nabla f(X_n) = \nabla f(X_{n+1})$, which would imply that a critical point has already been reached:
\begin{equation}
    h^{opt}_n = \max \left(0 , 2 h_n \frac{\langle \nabla f(X_n)-\nabla f(X_{n+1}),\nabla f(X_n) \rangle }{||\nabla f(X_n)-\nabla f(X_{n+1})||^2} \right).
\label{eq:hopt0}
\end{equation}
Note that it is important that in \eqref{eq:hopt0}
 the matrix $A$, which is impossible to compute, does not appear; only appear $\nabla f(X_n)$ and $\nabla f(X_{n+1})$
that are known.\footnote{Recall that do not discuss here the stochastic part; in practice an unbiased estimator of the 
gradient $\nabla f(X)$ is available.}.

To conclude: if the current learning rate is $h_n$ then it should be put to $ h^{opt}_n$ (given in equation \eqref{eq:hopt0}) to have the best convergence and stability properties.

Note that when $A$ is definite positive, the scalar product 
$p_n: =  \langle A\nabla f(X_n),\nabla f(X_n) \rangle$ must be positive. Therefore, if in a given iteration $n$, 
$\langle \nabla f(X_n) - \nabla f(X_{n+1}),\nabla f(X_n) \rangle$ (which equals $ h_n p_n$) happens to be negative
 this means that the second order assumption
\eqref{eq:forder2}
made on $f$ breaks down around  the current point $X_n$; we cannot trust any computation made above and thus when $p_n <0$ we can set for instance $h^{opt}_n = h_n$.
Therefore, denoting now $h_n$ the learning rate
at step $n$ we will define:
\begin{equation}
    h^{opt}_n = \left\{
    \begin{array}{ll}
         \ 2 h_n \frac {p_n}{|| \nabla f(X_n) - \nabla f(X_{n+1})||^2} & \mbox{if } p_n >0 \\
         \ h_n & \mbox{otherwise. }  \\
    \end{array}
\right.
\label{eq:hopt}
\end{equation}

\subsection{Update policy}

At a given iteration $n$, if $h_n \ll h^{opt}_n$, this means that the gradient descent is progressing too slow with the current learning rate and thus we need to accelerate it by increasing the learning rate. 
In practice, to not break the stability condition, the new learning rate must not be very close to  $h^{opt}_n$. To avoid this risk, we choose a gradual update with an hyper-parameter $\beta$ close to $1$:
\begin {equation}
    h_{n+1} = \beta h_n + (1 - \beta) h^{opt}_n.
\end {equation}
Suppose now that $h_n \gg h^{opt}_n$. This means that the current learning rate breaks the convergence criteria. Then, we have to decrease it; 
contrary to previous policy, here we do not want a slow update because instability is already set in. A drastic measure is required, otherwise the whole optimization may become useless. We propose the following update rule:
\begin {equation}
    h_{n+1} = (1 - \beta) h^{opt}_n.
\end {equation}
This update is not necessarily close to $h_n$ but is a conservative choice to enter again the stability region, which, in practice gives good results.

\subsection{The SGD-G2 algorithm}

We present in this section the algorithm resulting from the above considerations; due to the stochastic nature of the gradient that is to be computed, for each iteration $n$ in all the formulas in the former section the full gradient $\nabla f$ must be replaced by $\nabla f_{\gamma_n}$ . 
Then, our suggested adaptive SGD called "SGD-G2" is described by the Algorithm~\ref{algo:1} which includes the provision for mini-batch processing.

\begin{algorithm}[htb!]
   \caption{SGD-G2}
   \label{alg:sgd-g2}
\begin{algorithmic}
   \STATE {\bfseries Set hyper-parameter:} $\beta$, 
mini-batch size $M$,   choose stopping criterion
   \STATE {\bfseries Input:}  initial learning rate $h_0$, initial guess $X_0$
   \STATE {\bfseries Initialize iteration counter:} $n=0$ 
   \WHILE{ stopping criterion not met}
   \STATE select next mini-batch $\gamma_n^m$, $m=1,..., M$
   \STATE Compute $g_n= \frac{1}{M}\sum_{m=1}^M \nabla f_{\gamma _n^m} (X_n)  $
   \STATE Compute $
   \tilde{g}_n=  \frac{1}{M} \sum_{m=1}^M \nabla f_{\gamma _n^m} (X_n - h_n g_n) $
   \STATE Compute 
$$    h^{opt}_n = \left\{
\!\!\!    \begin{array}{ll}
          \frac {2 h_n \langle g_n -\tilde{g}_n,g_n \rangle}{\| g_n - \tilde{g}_n\|^2} & \mbox{if } \langle g_n -\tilde{g}_n,g_n \rangle >0 \\
         h_n & \mbox{otherwise. }  \\
    \end{array}
\right.$$
   \IF {$h^{opt}_n \ge h_n$}
   \STATE $h_{n+1} = \beta h_n + (1 - \beta) h^{opt}_n$
   \ELSE 
   \STATE $h_{n+1} =  (1 - \beta) h^{opt}_n$
   \ENDIF 
   \STATE Update $X_{n+1} = X_n - h_{n+1} g_n $
   \STATE Update $n \to n+1$
   \ENDWHILE
\end{algorithmic}
\label{algo:1}
\end{algorithm}

\begin{figure}[htb!]
\vskip 0.2in
\begin{center}
\centerline{\includegraphics[width=.95\columnwidth]{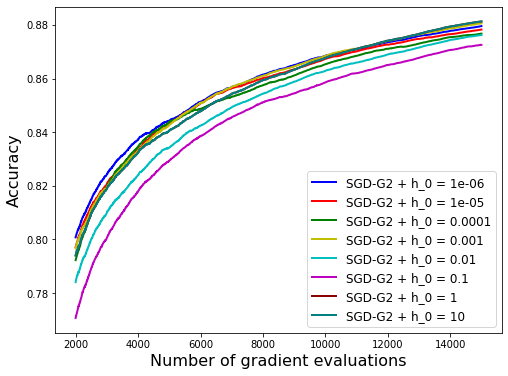}}
\caption{Numerical results for the SGD-G2 algorithm on the  FMNIST database with several choices of the initial learning rate $h_0$. Here $\beta=0.9$. Similar results are obtained for the MNIST and CIFAR10 databases.}
\label{fig:fmnistbeta}
\end{center}
\vskip -0.2in
\end{figure}

\begin{remark}
Several remarks are in order:
\begin{enumerate}
    \item The computation of both $g_n$ and $\tilde{g}_n$ allows in principle to construct a more precise, second order in the learning rate, estimate of the next step $X_{n+1}$
    of the form $X_n - \frac{h_n}{2}(g_n + \tilde{g}_n)$; this is not what we want here, the precise estimate is only used to calibrate the learning rate, in the end the SGD update formula is invoked to advance the network parameters $X_n$ to $X_{n+1}$.
\item It is crucial to have the same randomness in the computation of $\tilde{g}_n$
as the one present in the computation of $g_n$. This ensures that a consistent approximation is obtained as detailed in Theorem \ref{thm:cvsgdrk2}.
\item We recommend to take the initial learning rate $h_0$ very small, in order to be sure to start in the  stability region around $X_0$, for instance $h_0=10^{-6}$; however numerical experiments seem to be largely insensitive to this value as detailed in section \ref{sec:numerics}, figure \ref{fig:fmnistbeta}.
\end{enumerate}
\end{remark}

\section{Numerical experiments}
\label{sec:numerics}

\begin{figure}[ht]
\vskip 0.2in
\begin{center}
\centerline{\includegraphics[width=.95\columnwidth]{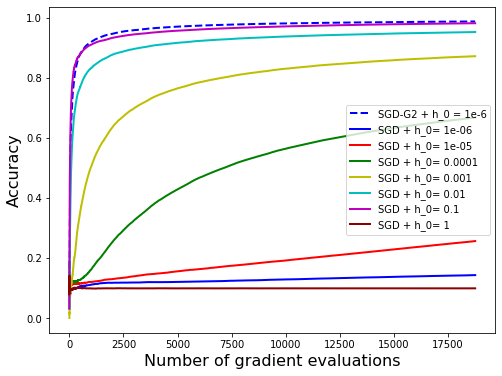}}
\caption{Numerical results for the SGD and SGD-G2 algorithms on the  MNIST database. Here $\beta=0.9$.}
\label{fig:mnist}
\end{center}
\vskip -0.2in
\end{figure}

\begin{figure}[htb!]
\vskip 0.2in
\begin{center}
\centerline{\includegraphics[width=.95\columnwidth]{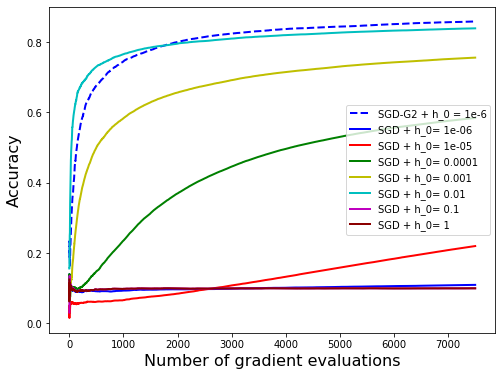}}
\caption{Numerical results for the SGD and SGD-G2 algorithms on the  FMNIST database. Here $\beta=0.9$.}
\label{fig:fmnist}
\end{center}
\vskip -0.2in
\end{figure}

\begin{figure}[ht!]
\vskip 0.2in
\begin{center}
\centerline{\includegraphics[width=.95\columnwidth]{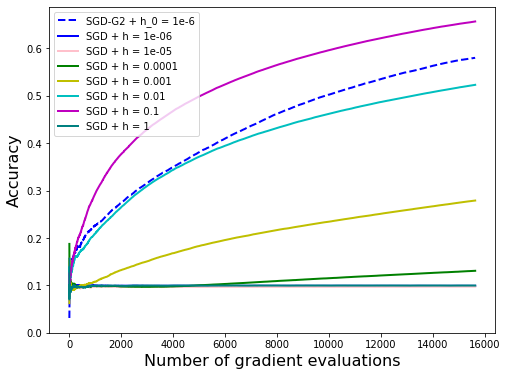}}
\caption{Numerical results for the SGD and SGD-G2 algorithms on the  CIFAR10 database. Here $\beta=0.9$.}
\label{fig:cifar10}
\end{center}
\vskip -0.2in
\end{figure}

\begin{figure}[ht!]
\vskip 0.2in
\begin{center}
\centerline{\includegraphics[width=.95\columnwidth]{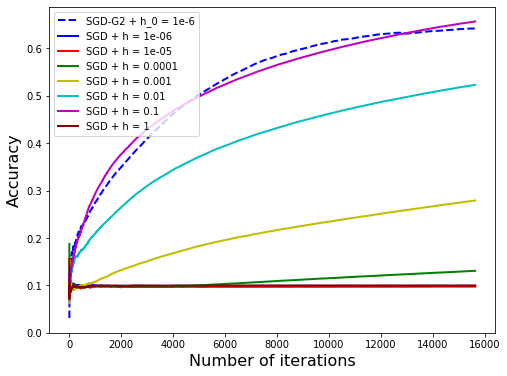}}
\caption{Numerical results for the SGD and SGD-G2 algorithms on the  CIFAR10 database. Here $\beta=0.9$, and we compare in number of iterations instead of gradient evaluations.}
\label{fig:cifar10iterations}
\end{center}
\vskip -0.2in
\end{figure}

\begin{figure}[htb!]
\vskip 0.2in
\begin{center}
\centerline{\includegraphics[width=.95\columnwidth]{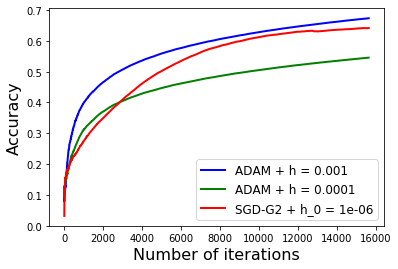}}
\caption{Numerical results for the SGD-G2 algorithm and the ADAM algorithm on the  CIFAR10 database. Here, $\beta = 0.9$.}
\label{fig:cifar_sgd_adam1}
\end{center}
\vskip -0.2in
\end{figure}


\subsection{Network architecture} \label{sec:architecture}

We conducted experiments on three different data sets (MNIST, Fashion MNIST and CIFAR-10) using neural networks (CNNs) developed for image classification. 
We follow in this section the specifications in 
 \cite{hidaka_cifar,tensorflow_cnn_tutorial} and in \cite{chen_mnist} and reproduce below the corresponding architectures as
given in the reference:

\noindent {\bf MNIST/Fashion-MNIST} ($28\times28$ sized images):  
three dense 256 neurons feed-forward ReLU layers 
followed by a final dense 10 neurons feed-forward layer.

\noindent {\bf CIFAR-10} ($32\times32$ images with $3$ color layers):
a convolution layer with $3\times3$ filters,  a $2\times 2$ max pooling layer, 
a convolution layer with $3\times3$ filters,  a $2\times 2$ max pooling layer, 
a convolution layer with $3\times3$ filters followed by a flatten layer, a dense layer with a $64$ neurons and a final dense layer with $10$ neurons.
All activations are ReLU except last one which is a softmax.

The last layer returns the classification result.

All hyper-parameters are chosen as in the references: the loss was minimized with the SGD / SGD-G2 algorithms and hyper-parameters $\beta=0.9$ or as indicated in the figures; we used $10$ epochs. The batch size is $32$.

\subsection{Discussion}
 
First, we compare the 
performance of the adaptive SGD algorithm for different choices of the $h_0$ parameter. The results do not vary much among databases, we plot in figure \ref{fig:fmnistbeta} the ones for FMNIST. The variability is similar for MNIST but slightly larger for CIFAR10.

We come next to the heart of the procedure and we compare the performance of the adaptive SGD algorithm with the standard SGD algorithm for different initial learning rates.
Recall that the goal of the adaptive procedure is not to beat the best possible SGD convergence but to identify fast enough the optimal learning rate. We recommend thus to start from a very small value of $h_0$; the algorithm will increase it up to the stability 
threshold. This is indeed what happens, see figures \ref{fig:mnist}, \ref{fig:fmnist} and \ref{fig:cifar10}.
As the adaptive algorithm uses two (mini-batch) gradient evaluations per iteration, we take as $x$-axis in the plots the number of gradient evaluations and not the iteration counter, (in order not to favor the adaptive procedure which is more costly per iteration). We see that in all situations, starting from a tiny value of $h_0$, the SGD-G2 algorithm quickly reaches the stability region and converge accordingly. 

For MNIST and FMNIST the SGD-G2 cannot be surpassed by SGD, even for the optimal SGD learning rate; on the contrary for CIFAR10 the SGD with the optimal learning rate (which has to be searched through repeated runs) does converge better than the SGD-G2, but this is due to the counting procedure: if instead of number of gradient evaluations we count the iterations, the two are comparable as shown in figure \ref{fig:cifar10iterations} (same considerations apply for the Adam algorithm as illustrated in figure \ref{fig:cifar_sgd_adam1}); so one can imagine that the adaptive part is only switched on from time to time (for instance once every $10$ iterations), which will make its overhead negligible and still reach the optimal learning rate regime. Such fine tuning remains for future work.

In conclusion, an adaptive SGD algorithm (called SGD-G2) is proposed which is both computationally convenient and provides a stable, optimal learning rate value. The procedure is tested successfully on three image databases (MNIST, FMNIST, CIFAR10).


\bibliography{rk2_refs}
\bibliographystyle{plain}
\end{document}